# A Model for Spatial Outlier Detection Based on Weighted Neighborhood Relationship

Ayman Taha    Hoda M.Onsi    Mohammed Nour El din    and Osman M. Hegazy
Faculty of computers and information –Cairo University

**Abstract:** Spatial outliers are used to discover inconsistent objects producing implicit, hidden, and interesting knowledge, which has an effective role in decision-making process. In this paper, we propose a model to redefine the spatial neighborhood relationship by considering weights of the most effective parameters of neighboring objects in a given spatial data set. The spatial parameters, which are taken into our consideration, are distance, cost, and number of direct connections between neighboring objects. This model is adaptable to be applied on polygonal objects. The proposed model is applied to a GIS system supporting literacy project in Fayoum governorate.

1- Introduction

Outlier detection is one of the most famous techniques in data mining. The objective of this technique is to discover a set of data objects, which are abnormal, or unpredicted and inconsistent with the general behavior of neighboring data objects. These objects may be considered as noise or error with respect to the remainder of data objects. Therefore many data mining techniques ignore, eliminate or minimize their influence [1]. But other techniques consider the analysis of these objects to be important since they may carry interesting, unexpected, and implicit knowledge. Outlier Analysis can be beneficially used in many applications such as credit card fraud, voting irregularity, and weather prediction [2], [3], and [4].

The nature of spatial dataset differs from traditional databases since the objects' behaviors in a spatial database are dependent on their neighbor behaviors. The weight of dependence is inversely proportional to the distance between these objects [5]. Whereas the traditional database builds on the fact that data samples are independently generated. Therefore, the techniques and algorithms of traditional data mining are altered to take into consideration this property, which is called spatial autocorrelation [6], [7], and [8].

In spatial domain, we can define spatial outliers as a set of data objects that are displayed as abnormal, unexpected and inconsistent with neighboring objects which have an effect on the data objects through a predefined spatial relationship [9] and [10]. Spatial relationship is based on a set of spatial factors such as distance, cost, number of direct connections, and relative accessibility between objects… etc. Spatial outliers can be useful in many areas such as transportation, ecology, and location based services [11] and [12].

In this paper, we propose a new framework to define neighboring objects. We change the form of neighborhood relationships by adding a new argument, which considers weighted effects of neighboring objects on any specified object. We use more than one spatial attribute to define neighboring objects, namely; distance, cost, and number of direct connections between the objects. We provide a model to calculate this weight. We also adapt the correlated spatial outlier detection algorithm to meet the changes in neighborhood relationship due to weight consideration. Finally, we extend this model to be applied to polygons and hence, we apply the proposed model into Fayoum governorate project for Female Literacy Program.

**2- Related Work**

In outlier detection algorithms, we should distinguish between two main sets of attributes; one set of attributes is used for defining neighborhood relationships, and the other set is used for comparison of an object with its neighbors to detect spatial outliers. As shown in Figure (1); classical researches in spatial data mining classify outliers in spatial datasets into two categories; One-dimension and Multi-dimension.

The first category, One Dimension, uses only one of the non-spatial attributes for comparison purposes. In this case, the neighbor's features are ignored which means ignoring spatial attributes and spatial relationships between objects. On the other hand, multi-dimension category uses some spatial or non-spatial attributes for defining neighbors and others for comparison purposes. Multi-dimension outlier detection techniques may be further divided into two subgroups, namely, homogeneous dimensions, and spatial outliers.

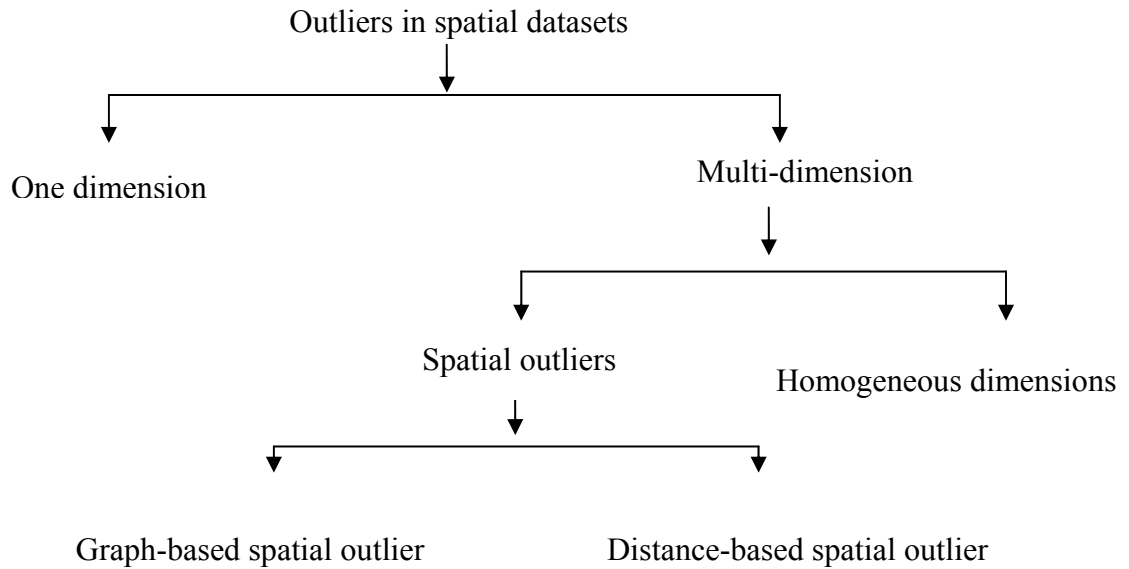

Figure (1)   Classification of Outlier Detection Techniques in Spatial Dataset

Homogeneous dimensions techniques use spatial such as distance between two villages and non-spatial attributes such as number of populations in each village for defining neighbors and other spatial and non-spatial attributes for comparison purposes. There is no separation between spatial and non-spatial attributes. Many outlier detection methods were recently proposed in homogeneous dimensions outliers taking into consideration distance as a measure of outliers such as Distance-based outliers [13], [14], [15], and [16]. In [14], and [15], the term distance-based outliers is used and defined as an object O in dataset T to be distance-based outlier DB (p,d)-outlier, where p is a fraction of all objects in T that lie within a distance greater than d from this object. In [16] the authors find the K-nearest object for each object in a dataset, and then they rank all objects descendingly by the K-nearest neighbor distance, and then they define the top N objects as outliers. But in [13] the data is stored in clusters. In each cluster, outlier detections tests have been done independently from other clusters. The resulting outliers are called local outliers because they are relative to each cluster.

These methods have many disadvantages namely, they do not distinguish between spatial and non-spatial attributes, and they detect global outliers not spatial outliers (general outliers mean objects that have high difference in their behaviors from the normal behavior of others objects in the space, but spatial outliers mean objects that have high difference in their non-spatial behaviors from the normal behavior of their spatial neighbors) [17].

On the other hand, spatial outliers' algorithms use both spatial attributes and non-spatial attributes. They use spatial attributes for defining neighborhood relationships and non-spatial attributes for comparison purposes [7]. Spatial outlier detection techniques can be further divided into two subcategories; graph-based spatial outliers and distance-based spatial outliers. The difference between them lies in the definition of spatial neighborhood relationships as shown in Figure (2). In graph-based spatial outliers, the set of spatial neighboring objects are defined as all objects that have direct connections to a specified object. But distance-based spatial outliers' techniques calculate spatial neighborhood relationships based on distance by making a buffer zone around each object to discover its neighbors [8]. Figure (2-a) shows the input spatial data set T that contains a set of objects {A, B, C, D, E, F, G, H, J, K, L}, direct connections between these objects, and the length of theses connections. We now want to define the neighboring objects of an object A. Figure (2-b) shows the distance-based neighboring objects of A which include all objects inside the buffer zone which includes {B,K,J,H}. But if we want to define graph-based neighbors, the set of neighboring objects will include all objects that are directly connected to A {B, D, E} as indicated in Figure (2-c).

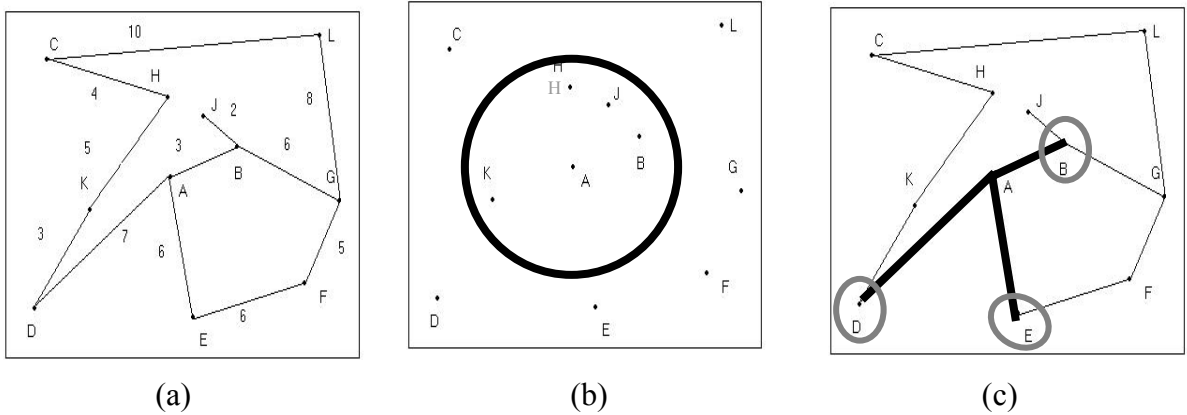

(a)        (b)        (c)

Figure (2)   Comparison of Different Spatial Neighbors.
(a) Spatial Data Set      (b) Distance Based Neighbors of Node A
(c) Graph Based Neighbors of Node A

Although a large amount of algorithms try to discover spatial outliers, they have many limitations. First, most of spatial outlier algorithms assume that all neighbors to a certain object have the same weight of effect on this object, but the nearest object should have the largest weight of effect according to the first law of geography [5]. Second, there is no standard to define the spatial neighborhood relationships; some algorithms leave these relationships to be defined manually by users. Other algorithms use external software to define them. This leads to different results for the same application using the same factors. Third, classical researches separate between graph-based spatial neighbors and distance-based spatial neighbors. But in many realistic applications, we should use connections and distances to discover spatial outliers. Last, such algorithms are designed to be applied only to points and no extensions of these algorithms could be applied to more complex objects such as lines or polygons.

**3- The Proposed Model**

The most important limitation of classical techniques in spatial outlier detection is the assumption, which supposes that all neighboring objects have the same effect on a specified object, which is not realistic. In fact each neighbor object has its own weighted influence on the specified object. Applying this concept on distance-based spatial outliers algorithms, the nearest object has the largest effect. For graph-based spatial outliers algorithms, the objects that have direct connections has larger effect than that object which has indirect connection. Also, the object that has the smallest cost with respect to a specified object has the largest effect on it.

So we suggest to change the form of spatial neighborhood relationship from (X,Y) to (X, Y, $weight_{XY}$) which means that an object Y is affected by an object X, and $weight_{XY}$ is the quantity of effect. This quantity must be between 0 and 1. The sum of weights that affect the same object must be equal to 1. We should distinguish between $weight_{XY}$ and $weight_{YX}$. From this parameter, we define the objects that have the largest and the smallest effects on a specified object. Hence, we should change the spatial outlier detection algorithms that use normal average to use weighted average. This form can be deduced from bilinear interpolation.

If we have an object r, which has N neighbors object, then

$$E(r) = \sum F(i) / N \qquad \text{For } i=1,\ldots N \qquad (1)$$

Where $F(i)$, $E(r)$ are the actual attribute value at object i, and the expected attribute value at object r respectively, and N is the number of neighboring objects relative to object r.

According to our proposed model equation (1) is modified to equation (2) as follows:

$$E(r) = \sum W_{ri} F(i) \quad \text{For } i=1,\ldots N \quad (2)$$

Where $W_{ri}$ is the weight of the effect from point i on point r, and the summation of $W_{ri}$ must be 1.

The problem reduces to how to calculate this weight automatically. We also propose here a simple method to calculate this weight through a single spatial factor, namely, the distance between the objects. Then we extend this method to cover many spatial factors, controlling the influence of each. Since the distance is inversely proportional to the weight of the effect, the longest distance between objects indicates the lowest effect between them as indicated by equation (3)

$$W_{ri} = (1/D_{ir}) / \sum (1/D_{ir}) \quad \text{For } i=1,\ldots N \quad (3)$$

Where $D_{ir}$ is the distance between the two points i,r.

Considering the effect of $R_{ir}$, the number of direct connections between the two points i and r, the largest value of $R_{ir}$ indicates the easiest movement from i to r. The corresponding weight can be expressed as follows:

$$W_{ri} = (R_{ir}) / \sum (R_{ir}) \quad \text{For } i=1,\ldots N \quad (4)$$

Taking into consideration the two previous factors together, we have:

$$W_{ri} = (\alpha / D_{ir}) / \sum (1/D_{ir}) + (1-\alpha)(R_{ir}) / \sum (R_{ir}) \quad \text{For } i=1,\ldots N \quad (5)$$

Where $\alpha$ is the weight of the effect of distance which must be $0 < \alpha < 1$.

Additionally, the cost factor to move from point i to point r, $C_{ir}$, is inversely proportional to the weight. Thus, the overall weight can be formulated as:

$$W_{ri} = (\alpha / D_{ir}) / \sum (1/D_{ir}) + (\beta R_{ir}) / \sum (R_{ir}) + (\delta / C_{ir}) / \sum (1/C_{ir})$$
$$\text{For } i=1,\ldots N \quad (6)$$

Where α, β and δ are coefficients to distance, number of direct connections and cost, respectively, where   α + β + δ = 1.

Equation (6) merges the most common three factors in point data types, which are distance, number of direct connections and minimal cost.
This way we eliminate the separation between distance-based spatial outliers and graph-based spatial outliers, which is found in literature.
We provide methods to calculate these factors automatically. These methods are:
    i) Distance, by making a buffer zone around each object with a radius r defined by the user, which means neglecting the influence of the distance of objects outside this area. All objects inside this area are considered neighbors to this object.

    ii) Number of direct connections, can be calculated by finding the number of direct links passing through this object.

    iii) Minimal cost between two points, by defining the number of direct and indirect connections between the two points, and then finding the minimal cost for these connections. We can also define a limit L for the cost factor. If the minimal cost is greater than L then we will ignore the relationship to this node.

**4- Spatial outlier Definition Adaptation**
After the suggested changes have been made through the proposed model, the definitions of spatial outliers and the family of algorithms provided by Shekhar [7], should be adapted to these changes.

Consider spatial framework SF = (S,NB), where S is a set of locations $\{s_1,s_2,s_3,\ldots,s_n\}$ and NB (S,S,W) where W is the weight of the effect and W $\in$ [0,1], NB is a neighbor relation over S. We define neighborhood of location x in S using NB as N (x), specifically N(x) = $\{y \mid y \in S, NB(x,y,W_{yx}), \text{ and } 1 > W_{yx} > 0\}$.

**Definition:** An object O is an S-outlier (f, $f^N_{aggr}$, $F_{diff}$, ST) if ST $\{F_{diff} [f(x), f^N_{aggr}(f(x),N(x))]\}$ is true, where f : S $\rightarrow$ R is an attribute function, $f^N_{aggr} : R^N \rightarrow R$ is an aggregate function for the values of f over neighborhood, R is a set of real numbers, $F_{diff}$ R x R $\rightarrow$ R is difference function, and ST : R { true , false }is statistical test procedure for determining statistical significance.

We can define f function as the non-spatial attribute, $f^N_{aggr} = E_{y \in N(x)}(f(Y))$ as the weighted average of attribute values over neighborhood $N(x)$, $F_{diff}(x)$ is called the difference function $S(x) = f(x) - f^N_{aggr}$ is the arithmetic difference between attribute function and neighborhood aggregate function $f^N_{aggr}$. Let $\mu_{S(x)}$ $\sigma_{S(x)}$ be the mean and standard deviation of the difference function $F_{diff}$, respectively.

Then the significance test function ST can be defined as $Z_{S(x)} = |S(x) - \mu_{S(x)} / \sigma_{S(x)}| > \Theta$, where $\Theta$ is the residual error. The choice of $\Theta$ specifies the confidence level. For example if 95% confidence level is required, $\Theta = 2$.

**5- Application**

Egyptian Center for Women's Rights (ECWR) in participation of Social Research Center (SRC) at American university in Cairo (AUC) started the National Project for Literacy that aims at decreasing the percent of illiteracy in the Egyptian villages especially among females. The project focuses on percent of illiterate females between 14 and 35 years old. They started with El-Fayoum governorate which contains 167 villages grouped around 6 cities. For each village, we have available the percentage of illiterates among females, males, and total population. Figure (3) shows the map of the study area. In our application, we are interested in discovering outliers' villages which have the percent of illiteracy in females that highly differs from their neighboring villages.

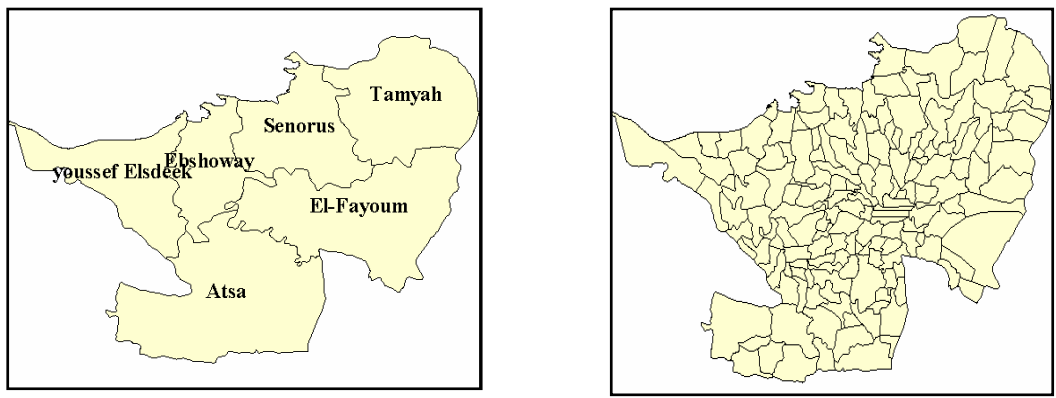

(a)            (b)
Figure (3) Snapshot Maps from El-Fayoum Literacy Project.
(a) Map of El-Fayoum Cities     (b) Map of El-Fayoum Villages.

To apply the proposed model in this application, some limitations have been considered. First, the data type of objects is polygons not points. Second, we need to define neighboring objects. Third, calculation of the weight effect between neighboring objects. Finally, calculation of the distance between two polygons.

We make the next assumptions to help in solving the problems:
- We define the spatial neighbors to a certain object as all adjacent objects which have a common line with that object.
- We use distance between objects and area of neighboring objects to calculate the weighted effect, the distance being inversely proportional to the weighted effect, while the area being directly proportional to the weighted effect.
- We consider the distance between objects as the distance between their centers.

We compare between the results of our proposed model and those of applying classical spatial outlier algorithms [7]. As an example, in our model, we have village_Id 27 having 7 neighboring villages as shown in Figure (4). These neighboring villages have different weighted effect values on village_Id 27. We found that village_Id 29 is the nearest village, so it has the largest value (nearly 41% of the total effect). Whereas village_Id 42 has the smallest value (about 5%). The classical model assumes that the two villages (village_Id 29 and village_Id 42) have the same weight of effect (about 14,3%). This difference made our expected percent of illiterate females in this village 28%, while it is 45% in the classical model. We already have the actual percent of illustrate females as 26%.

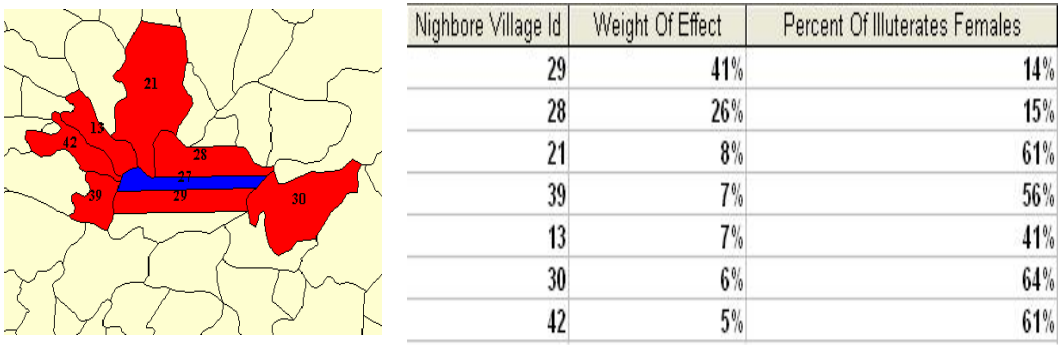

(a)  (b)

Figure (4) Samples From The Result of the Proposed Spatial Outliers Detection Algorithm In El-Fayoum Literacy Project.
(a) Village_Id 27 and Its Neighboring Villages
(b) Table of Weight of Effect and Percent of Illiterate Females for Each Neighboring Village Ordered According to Its Effect

The difference error S(x) between our proposed model and the classical model can be calculated by equations (7) and (8)

S(x): 28% - 26% = 2%     for our proposed model     (7)

S(x): 45% - 26% = 19%     for the classical model     (8)

We compare between our proposed model and the classical model by the mean square measure (neglecting the sign of the difference error). We found that our proposed model, in some cases, decreases the difference error by 98% as indicated by equations 9, 10. We also found that the proposed model decreases the difference error by 8% on average.
The difference between the square of difference error of the classical model and the proposed model is:

$$(0.19)^2 - (0.02)^2 = 0.0357 \qquad (9)$$

The percentage of improvement which is made by the proposed model in this village is:

$$(0.19)^2 - (0.02)^2 / (0.19)^2 = 98.89\% \qquad (10)$$

In this application, we need 95% confidence level, so we take $\Theta = 2$ so the spatial outlier object has $| S(x) - \mu S_{(x)} / \sigma_{S(x)} | > 2$. Samples of results of significance test function ST of our proposed model and the classical model are shown in Figure (5). The set of spatial outliers villages in the proposed model contains {17, 216, 238, 26, 317, 511, 302, 239, 30}, but in the classical model contains {17, 216, 238, 26, 317, 28, 29, 30}. If we compare the results of the two models, we find that the next objects {17, 216, 238, 26, 317, 30} are spatial outliers in the two models. But the proposed model detects {511, 302, 239} as outliers, while the classical model does not detect them, and vice versa. The classical model detects {28, 29} as an outlier, but the proposed model does not detect it as outlier because the most nearest villages to this village have similar percentage of female illiterates.

| Village_ID | Proposed Model | Classical Model |
|---|---|---|
| 216 | -2.74 | -2.61 |
| 17 | -2.70 | -2.59 |
| 238 | -2.46 | -2.51 |
| 26 | -2.29 | -2.25 |
| 317 | -2.28 | -2.10 |
| 29 | -1.87 | -2.32 |
| 28 | -1.76 | -2.44 |
| 511 | 2.02 | 1.88 |
| 302 | 2.04 | 1.68 |
| 239 | 2.12 | 1.96 |
| 30 | 2.57 | 2.52 |

Figure (5)   Samples of Results of Significance Test (ST) Function of Our Proposed Model and the Classical Model.

## 6- Conclusions

In this paper, we proposed a new definition to spatial neighborhood relationships adding a new parameter which is called weight of effect. We also changed the definition of spatial outliers to accommodate the new definition of spatial neighborhood relationships. We analyzed the classical outlier detection methods in spatial data set merging the graph-based and distance-based spatial outliers. We proposed a new model to detect spatial outliers objects based on many spatial features such as distance, number of direct connections, and cost. We extended the proposed model to be applied to more complex data types such as polygonal objects. In addition, we provided experimental results from the application of the proposed model on El-Fayoum Literacy Project. Experimental results proved that our model decreases the error that results from the difference between the calculated value and the actual value by 8% on the average. This proposed model can be beneficially used in many applications such as transportation to discover congestion places, health and location based services.

## 7- References